\documentclass[runningheads]{llncs}

\usepackage{eccv}

\usepackage{eccvabbrv}

\usepackage{graphicx}
\usepackage{booktabs}

\usepackage[accsupp]{axessibility}  

\usepackage{multirow}
\usepackage[table,xcdraw]{xcolor}
\usepackage{algorithmic}
\usepackage{algorithm}
\usepackage[algo2e]{algorithm2e}
\usepackage{pifont}
\usepackage{subcaption} 

\usepackage[pagebackref,breaklinks,colorlinks,citecolor=eccvblue]{hyperref}

\usepackage{orcidlink}

\begin{document}

\title{EgoGVAE: Ego-body Mesh Reconstruction via Guided Variational Autoencoder} 

\titlerunning{EgoGVAE}

\author{Jaehun Jung\orcidlink{0009-0008-1815-5795} 
\and Wonjun Kim\thanks{Corresponding author}}

\authorrunning{J.~Jung and W.~Kim.}

\institute{Konkuk University \\
\email{\{brian111725, wonjkim\}@konkuk.ac.kr}
}

\maketitle

\begin{abstract}
  We address the problem of recovering the full-body mesh from only the head pose. This task has become essential for various applications based on head-mounted devices or smart glasses. 
  The challenge of this task lies in estimating the pose information of unobserved body parts based solely on a single joint (i.e., head) trajectory. 
  Several studies have begun to adopt head-conditioned generative models, however, such previous methods are costly and time-consuming due to the diffusion-based iterative process. 
  As an alternative, we propose a simple yet novel method that leverages the latent space of the guidance network, which is designed as a variational autoencoder taking full-body poses as inputs. 
  By enforcing latent distributions of this guidance network and our head-to-motion network to be similar, latent features sampled from the ‘guided’ distribution, i.e., distribution learned in our head-to-motion network, can be reliably decoded for natural representations of full-body poses even only with the head pose.
  One important advantage of the proposed method is that one-step sampling scheme achieves remarkably fast inference (more than 50 times faster) compared to diffusion-based approaches. 
  Experimental results on benchmark datasets show that the proposed method efficiently improves the performance of ego-body mesh reconstruction.
  The code and model are publicly available at: 
  \url{https://github.com/DCVL-3D/EgoGVAE_release}.
\end{abstract}

\section{Introduction}
\label{sec:intro}

As the era of physical AI approaches, interactive wearable devices are expected to become mainstream in our daily life.
At the same time, understanding 3D motions of a human wearer is essential for enabling diverse applications in such interactive environments.
For example, the algorithm implemented in the smart glass like Project Aria~\cite{engel2023project} needs to figure out the 3D status of the wearer based solely on SLAM head poses for assistive operations.
However, this task is quite challenging since most body parts of the wearer are unobservable in the egocentric input.

\begin{figure}[tb]
    \centering
    \includegraphics[height=3cm]{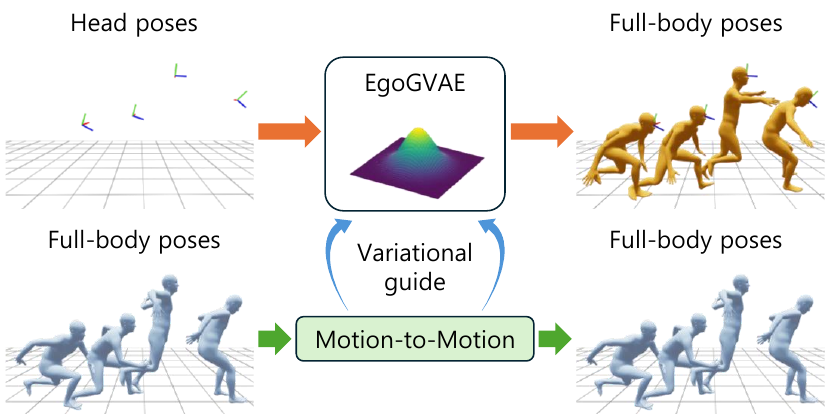}
    \vspace{-2mm}
    \caption{Overview of the proposed method that estimates full-body poses only from head poses via the guided variational autoencoder, i.e., EgoGVAE. Note that the motion-to-motion network guides our EgoGAVE in the latent space.
    }
    \label{fig:overview}
    \vspace{-2mm}
\end{figure}

In the beginning, to address this challenge, there have been meaningful attempts to utilize the hand position, which can be obtained from motion tracking sensors such as hand controllers and wristbands.
This explicit observation of the hand position helps the model predict the global orientation~\cite{jiang2022avatarposer} or joint-level features~\cite{zheng2023realistic}, but has the disadvantage of requiring additional hardware, leading to the limited accessibility in real-world environments.
Most recently, several studies have begun to adopt the diffusion model without using such extra hardware.
By taking the 3D head pose as the condition on learning the distribution of full-body poses, the prediction result of head-to-motion generation can be appropriately constrained.
In addition, normalization techniques of 3D head positions, e.g., fixing starting positions of motion sequences~\cite{li2023ego} or aligning per-timestep positions to the ground plane~\cite{yi2025estimating}, have been also applied to help train the diffusion model to focus on the intrinsic trajectory of full-body poses, regardless of the absolute 3D head position within the input sequence.
Even though diffusion-based approaches for head-to-motion generation have shown promising results, those require the expensive computational cost due to the iterative process and often result in unstable full-body poses when attempting to reduce the number of denoising steps.

In this paper, we propose a simple yet novel variational method, so-called EgoGVAE, that reconstructs the full-body mesh from only the head pose of the wearer.
The key idea of the proposed method is to leverage the latent space of the motion-to-motion network to guide the process of head-to-motion generation as shown in Fig.~\ref{fig:overview}.
Since the motion-to-motion network, which is designed to a transformer-based variational autoencoder, takes full-body poses as inputs, its latent distribution efficiently captures various characteristics of full-body poses and serves as a reliable prior for guiding the generation process.
By encouraging the latent distribution of our head-to-motion network to be similar to that of the guidance network, i.e., motion-to-motion network, 
latent features sampled from the learned distribution can be reliably decoded for natural representations of full-body poses even only with the head pose.
To align the generation process of the head-to-motion network with that of the guidance network, we concatenate the head pose with learnable tokens, which correspond to unobserved body parts. 
This appended embedding effectively mitigates the gap between latent spaces driven by two different inputs, i.e., head poses and full-body poses.
It is noteworthy that our guidance scheme is applied only to the training phase, leading to the fast inference with a one-step sampling compared to iterative diffusion-based approaches.
The main contribution of the proposed method can be summarized as follows:

\begin{itemize}
\item 
We propose to leverage the latent space of the guidance network, which takes full-body poses as inputs, for ego-body mesh reconstruction.
By enforcing latent distributions of this guidance network and our head-to-motion network to be similar, the model can easily understand the process of the head-to-motion generation and successfully generate full-body meshes with natural poses.

\item 
We represent the latent distribution of full-body poses in a variational framework. That is, the guidance process can be efficiently achieved by simply aligning Gaussian parameters estimated from head-to-motion and motion-to-motion networks, respectively. Moreover, this design scheme, which operates with one-step sampling in inference, makes the proposed method perform very fast (more than 50 times faster) compared to diffusion-based approaches.
\end{itemize}

\section{Related Work}

In this Section, we give a brief review of previous studies for 3D human mesh reconstruction (HMR) from third-person inputs and discuss representative approaches specifically designed to address challenges related to the egocentric input.

\subsection{3D Human Mesh Reconstruction}
The majority in the field of 3D human mesh reconstruction has utilized third-person inputs. 
The early methods adopted a simple encoder-decoder architecture to directly regress pose and shape parameters of the SMPL model~\cite{loper2023smpl}.
For example, Kanazawa \textit{et al.}~\cite{kanazawa2018end}
used the end-to-end convolutional neural network with the adversarial loss to achieve the plausible mesh reconstruction.
Kolotouros \textit{et al.}~\cite{kolotouros2019learning}
designed the optimization loop that enables the regression model to be trained with the strong supervision, which significantly improved the performance of human mesh reconstruction.
However, in-the-wild images often contain unobservable body parts occluded by objects, other people, or body parts of the target subject.
To improve the prediction accuracy under such occlusions,
Kocabas \textit{et al.}~\cite{kocabas2021pare}
proposed a part attention module that considers dependencies of visible body parts.
Sun \textit{et al.}~\cite{sun2021monocular}
predicted body center positions of every person in a full frame, which are then used to accurately estimate SMPL parameters for multiple people.
They also extended their work by explicitly considering the relative depth of each person through the bird’s-eye-view body center heatmap~\cite{sun2022putting}.
Dwivedi \textit{et al.}~\cite{dwivedi2024tokenhmr}
utilized the discrete pose codebook as strong pose priors to enforce plausible mesh reconstructions for occluded body parts.
Gwon \textit{et al.}~\cite{gwon2025eigenpose} proposed a new concept of eigenpose to understand various occlusions in a global context.
In addition, diverse methods have been introduced to further improve the performance of HMR in videos.
Specifically, Ye \textit{et al.}~\cite{ye2023decoupling}
applied both relative camera motions and tracking results to the joint optimization framework, which yields trajectories of the moving people in the world coordinate. 
Shin \textit{et al.}~\cite{shin2024wham}
utilized 2D keypoints and camera angular velocities to estimate global human trajectories while maintaining the temporal consistency in videos.
Most recently, Wang \textit{et al.}~\cite{wang2025prompthmr}
designed a multi-modal prompt encoder to refine the mesh reconstruction conditioned on the spatio-temporal information, which demonstrates the significant improvement in accuracy of HMR.

\subsection{Ego-body Mesh Reconstruction}
Although many methods have been explored for 3D human mesh reconstruction, the problem of estimating the 3D model from the egocentric input is still challenging due to the lack of visual clues for body parts of the target subject. 
To tackle this challenge, there have been impactful attempts to utilize explicit observations of the hand position.
Specifically, Jiang \textit{et al.}~\cite{jiang2022avatarposer}
used the trajectory of the head and two hands to estimate the global orientation, which serves as the root transform for the forward kinematic chain of the human body.
Zheng \textit{et al.}~\cite{zheng2023realistic}
also utilized such explicit observations to generate the initial joint-level features, which are calibrated through spatio-temporal transformers and subsequently fed into the SMPL regressor.
Du \textit{et al.}~\cite{du2023avatars}
designed a simple diffusion model to generate full-body poses with sparse IMU tracking signals in real-time. 
Feng \textit{et al.}~\cite{feng2024stratified}
first predicted upper-body poses by using positions of the head and two hands and then leveraged these prediction results as the condition for the diffusion-based generation process of lower-body poses. 
On the one hand, there have been several attempts to utilize visible hand observations when they appear in the field of view based on head-mounted cameras~\cite{aliakbarian2023hmd, jiang2024egoposer, chi2024estimating, barquero2025sparse}. 
For example, Jiang \textit{et al.}~\cite{jiang2024egoposer} 
proposed to exploit intermittent hand positions and orientations only when inside a headset’s field of view.
Barquero \textit{et al.}~\cite{barquero2025sparse}
proposed a rolling prediction model that leverages previously generated past motions and forecasted future motions to enable smooth transitions when hand positions are lost and later recovered.

On the other hand, to avoid unnecessary constraints in terms of using hand signals, a few studies have predicted full-body poses based solely on head poses.
Li \textit{et al.}~\cite{li2023ego}
extracted the 3D head pose by refining SLAM results with the gravity direction and optical flow features, and injected it into the diffusion model to generate full-body poses.
Similarly, Yi \textit{et al.}~\cite{yi2025estimating} 
also utilized the diffusion model, but assumed that head poses are directly computed from SLAM systems introduced in Project Aria~\cite{engel2023project}.
They mapped these poses onto the floor plane at each timestep to achieve both spatial and temporal invariance, helping the model train more robustly.
While previous approaches have adopted diffusion models to generate full-body poses conditioned on head poses, we shift our focus to the efficient variational method that leverages the generation process learned from full-body poses as guidance for reliable ego-body mesh reconstruction.

\begin{figure*}[tb]
    \centering
    \centerline{\includegraphics[width=1\textwidth]{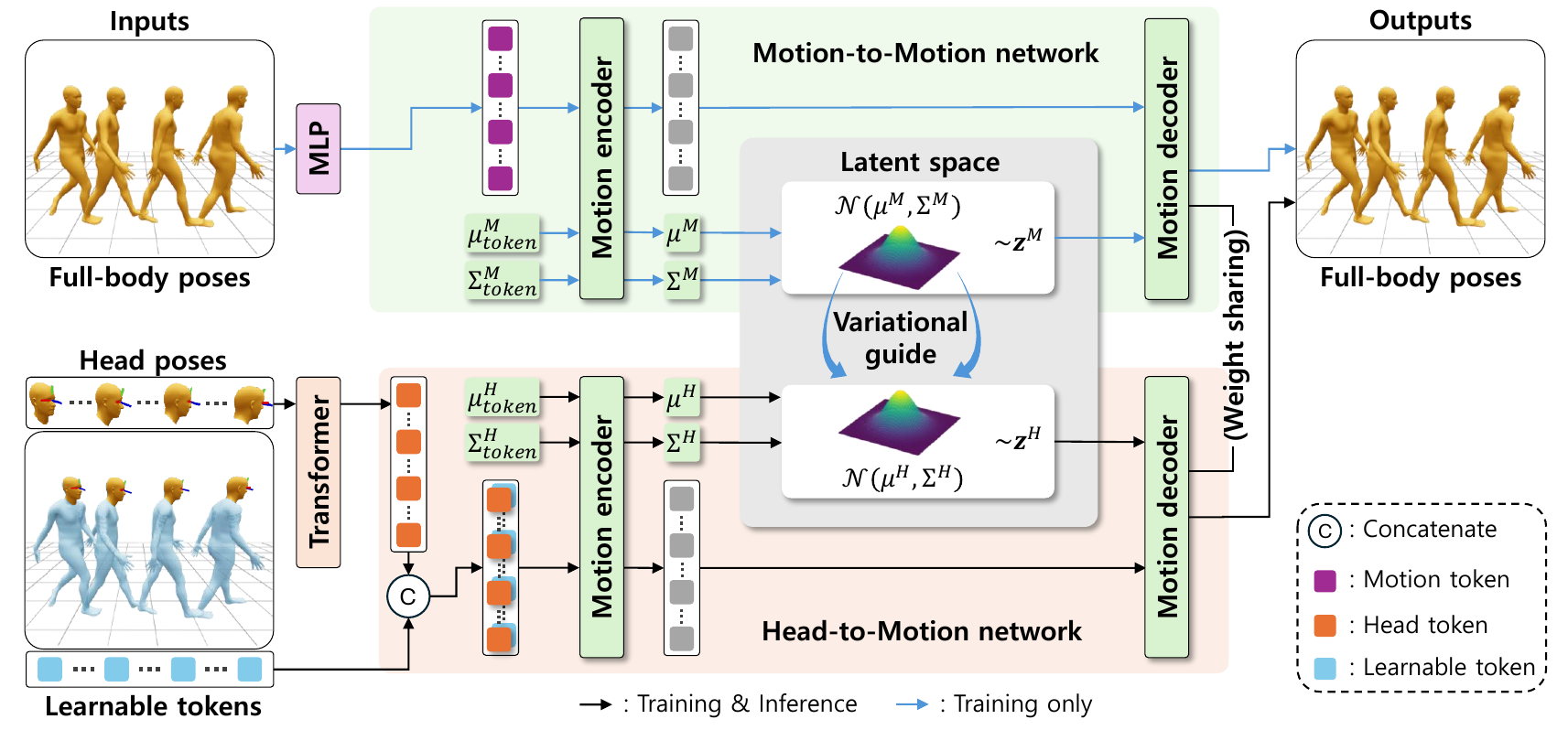}}
    \caption{The overall architecture of the proposed method for recovering full-body poses from head poses. A motion-to-motion network is used to guide the generation process of the head-to-motion network in the latent space. Note that both tokens for mean and variance are appended to input tokens of each network to formulate latent distributions as Gaussian distributions in a variational framework.
    }
    \label{fig:overall} 
    \vspace{-2mm}
\end{figure*}

\section{Proposed Method}
\label{sec:method}
The proposed method consists of two main networks, i.e., head-to-motion network and motion-to-motion network, both formulated as variational models. 
The core of the proposed method is to leverage the latent space of the motion-to-motion network for guiding the generation process of the head-to-motion network.
The overall architecture of the proposed method is illustrated in Fig.~\ref{fig:overall}.

\subsection{Task Definition}
\label{subsec:Definition}
The goal of this task is to reconstruct a sequence of full-body meshes from the head trajectory.
The input is defined as a sequence of \(T\) head poses, \(W_{1:T}=\{w_1, w_2, ..., w_T\}\). 
Here, \(w_t \in \mathbb{R}^{16}\) represents the head pose at timestep \textit{t}, which is concatenated by the relative motion, the canonicalized orientation, and the absolute height of the head position.
Note that we adopt the continuous 6D representation~\cite{zhou2019continuity} for rotational components.
These poses are defined in the world coordinate system and canonicalized to remove the effect of global rotation and translation by following the representation scheme used in~\cite{yi2025estimating}.
This normalization yields the head trajectory that reflects only the motion of the wearer independent of global movements.

We represent the full-body pose as a sequence of \(T\) timesteps, \(X_{1:T}=\{x_1, x_2, ..., x_T\}\).
At each timestep \textit{t}, we predict the 3D human model, i.e., \(x_t=\{\theta_t, \beta, \psi_t\}\).
Here, \(\theta_t \in \mathbb{R}^{21\times6}\) and \(\beta \in \mathbb{R}^{16}\)  denote local joint rotations and time-invariant shape parameters, respectively, of the SMPL-H model~\cite{loper2023smpl}.
\(\psi_t\ \in \mathbb{R}^{21}\) denotes the probability of the per-joint contact.
Note that the global root pose is not directly estimated.
Instead, it is recovered later through a global alignment step that uses the input head pose, as described in~\cite{yi2025estimating}.
Therefore, the core objective of this task is to learn the head-to-motion network that can predict the corresponding sequence of full-body poses \(X_{1:T}\) from the given head trajectory \(W_{1:T}\), thereby enabling the reconstruction of reliable full-body meshes.

\subsection{Head-to-Motion via Variational Guidance}
\label{subsec:Guidance}

The proposed motion-to-motion network models a latent distribution of realistic full-body poses, providing a variational guide for the process of head-to-motion generation.
To do this, we adopt the transformer-based variational autoencoder that encodes pose sequences into a Gaussian latent distribution in a similar way to~\cite{petrovich2022temos}.
Specifically, local joint rotations \(\theta_t\) at each timestep are tokenized to form the sequence of pose features.
Then, mean token \(\boldsymbol{\mu}_{token}^M\) and variance token \(\boldsymbol{\Sigma}_{token}^M\) are appended to this input sequence as shown in Fig.~\ref{fig:overall}.
This appended input is processed by the transformer encoder, and the corresponding output embeddings are used to produce parameters \(\boldsymbol{\mu^M}\) and \(\boldsymbol{\Sigma^M}\) of the Gaussian distribution \(\mathcal{N}(\boldsymbol{\mu}^M,\boldsymbol{\Sigma}^M)\).
It is noteworthy that the motion-to-motion network learns the sequence-level latent distribution, which captures various characteristics of full-body poses and serves as the motion prior for the proposed framework.

The head-to-motion network mirrors the variational framework of the guidance network, i.e., motion-to-motion network, to construct its own latent distribution \(\mathcal{N}(\boldsymbol{\mu}^H,\boldsymbol{\Sigma}^H)\),
which is later aligned with \(\mathcal{N}(\boldsymbol{\mu}^M,\boldsymbol{\Sigma}^M)\).
Unlike the guidance network, the head-to-motion network is required to estimate the latent distribution of full-body poses from only the head trajectory \(W_{1:T}\).
To supplement the lack of visual clues, we introduce learnable tokens \(L_{token} \in \mathbb{R}^{T\times dim}\) corresponding to the rest of body parts, where \(T\) denotes the total number of timesteps.
More concretely, the input sequence for the encoder of the head-to-motion network is constructed in several steps as follows.
The head trajectory \(W_{1:T} \in \mathbb{R}^{T\times 16}\) is first transformed by a lightweight transformer to produce head embeddings \(E_W \in \mathbb{R}^{T\times dim}\).
Then, learnable tokens \(L_{token}\) are concatenated with these embeddings \(E_W\) and fed into the motion encoder (see Fig.~\ref{fig:overall}).
Consequently, the head-to-motion network aligns its generation process with that of the guidance network through learnable tokens.

\begin{figure*}[tb]
    \centering
    \centerline{\includegraphics[width=1\textwidth]{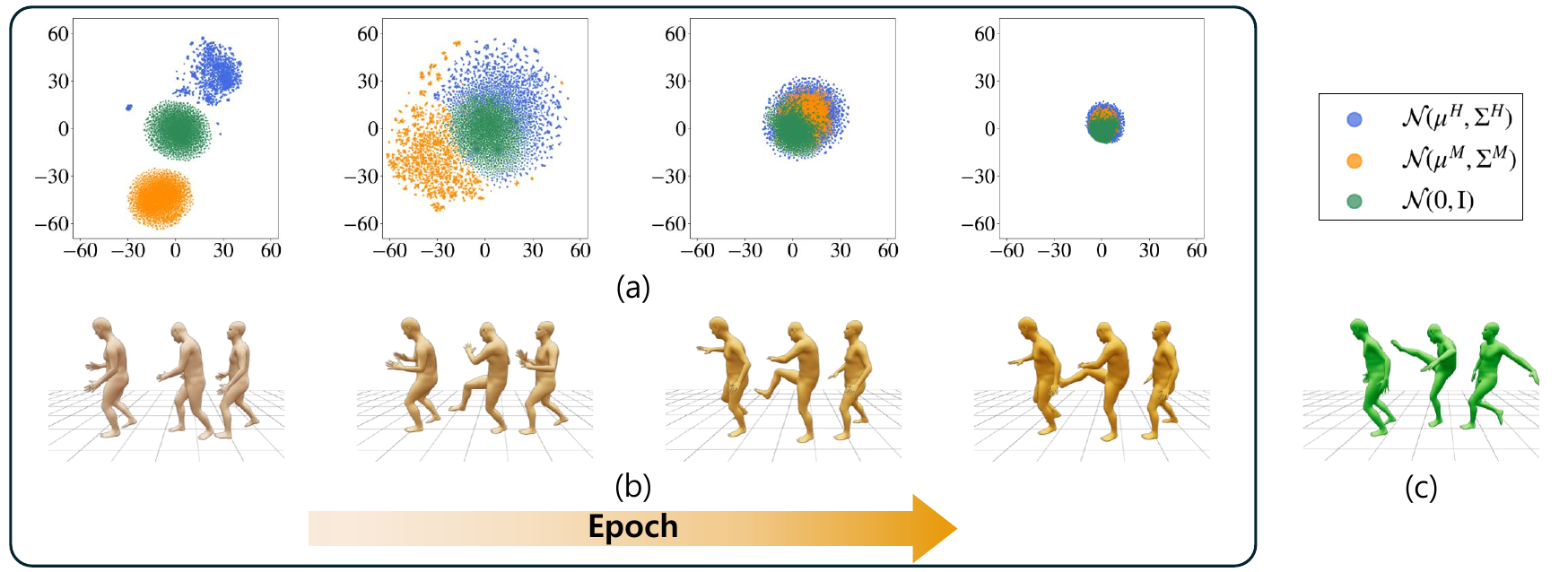}}
    \vspace{-1mm}
    \caption{Visualization for the alignment process of latent distributions.
    (a) The t-SNE~\cite{maaten2008visualizing} visualization of latent distributions over training epochs. 
    (b) Results of ego-body mesh reconstruction at corresponding epochs.
    (c) Ground Truth.
    }
    \label{fig:supp_dist} 
    \vspace{-2mm}
\end{figure*}

In order to enforce two latent distributions to be close to each other, we minimize the symmetric Kullback–Leibler divergence during training as follows:
\begin{equation}
\begin{aligned}
   \mathcal{L}^{H,M}_\text{KL}  &=  D_\text{KL}(\mathcal{N}(\boldsymbol{\mu}^H,\boldsymbol{\Sigma}^H) || \mathcal{N}(\boldsymbol{\mu}^M,\boldsymbol{\Sigma}^M)) \\
    &\quad + D_\text{KL}(\mathcal{N}(\boldsymbol{\mu}^M,\boldsymbol{\Sigma}^M) || \mathcal{N}(\boldsymbol{\mu}^H,\boldsymbol{\Sigma}^H)).
\end{aligned}
\label{eq:1}
\end{equation}
In addition, to regularize latent distributions of both networks, we encourage each distribution toward a standard normal distribution \(\mathcal{N}(\textbf{0},\textbf{I})\), which is defined as follows~\cite{petrovich2022temos}:
\begin{equation}
\begin{aligned}
   \mathcal{L}^{normal}_\text{KL}  =  D_\text{KL}(\mathcal{N}(\boldsymbol{\mu}^H,\boldsymbol{\Sigma}^H) || \mathcal{N}(\textbf{0},\textbf{I}))
    + D_\text{KL}(\mathcal{N}(\boldsymbol{\mu}^M,\boldsymbol{\Sigma}^M) || \mathcal{N}(\textbf{0},\textbf{I})).
\end{aligned}
\label{eq:2}
\end{equation}
Based on these two terms, the guidance is efficiently conducted in a variational scheme.
The effect of the proposed guidance scheme is shown in Fig.~\ref{fig:supp_dist}.
As can be seen, the latent distribution of the head-to-motion network $\mathcal{N}(\boldsymbol{\mu}^H,\boldsymbol{\Sigma}^H)$ is disjoint from that of the motion-to-motion network $\mathcal{N}(\boldsymbol{\mu}^M,\boldsymbol{\Sigma}^M)$ in the early stage of training (see the left side of Fig.~\ref{fig:supp_dist}(a)).
As the training epoch increases, these two distributions are progressively aligned to each other, and simultaneously converge to the normal distribution $\mathcal{N}(\textbf{0},\textbf{I})$ (see the right side of Fig.~\ref{fig:supp_dist}(a)).
Qualitative results of ego-body mesh reconstruction at the corresponding epoch are also shown in Fig.~\ref{fig:supp_dist}(b).
It is observed that full-body poses generated from head poses become more natural as latent distributions become aligned.

Along with this guidance, \(\textbf{z}^M\) and \(\textbf{z}^H\) are sampled from each distribution via the reparameterization trick~\cite{kingma2013auto}, and both are fed into the shared motion decoder, which is composed of transformer blocks~\cite{vaswani2017attention}, to generate full-body poses \(X_{1:T}\) during training.
Specifically, in the head-to-motion network, the decoder receives the latent feature \(\textbf{z}^H\) as the key and value, while the output of the motion encoder, i.e., concatenated embeddings of head and learnable tokens, is adopted as the query.
In contrast, in the motion-to-motion network, the decoder takes \(\textbf{z}^M\) as the key and value, and motion tokens encoded from full-body poses are used as the query to reconstruct \(X_{1:T}\).
Here, \textbf{z} represents the single latent feature of full-body poses over the entire sequence, while query tokens interact with \textbf{z} to decode this sequence-level information into poses at each timestep, maintaining temporal consistency with the input sequence.
At the inference time, a latent feature \(\textbf{z}^H\) is sampled in one step from the guided latent distribution \(\mathcal{N}(\boldsymbol{\mu}^H,\boldsymbol{\Sigma}^H)\) and subsequently decoded to reconstruct full-body meshes.
It is noteworthy that this process achieves the competitive performance at a remarkably fast inference speed compared to previous methods.

\subsection{Loss Function}
\label{subsec:loss}
Both the head-to-motion network and the motion-to-motion network are jointly trained based on three loss terms, i.e., reconstruction loss $\mathcal{L}_{\text{rec}}$, KL regularization $\mathcal{L}_{\text{KL}}$, and velocity loss $\mathcal{L}_{\text{vel}}$.
First, $\mathcal{L}_{\text{rec}}$ supervises both networks by comparing their predicted full-body poses $\hat{X}_{1:T}^H$ and $\hat{X}_{1:T}^M$ with the ground truth $X_{1:T}$.
Each predicted motion sequence $\hat{X}_{1:T}=\{\hat{x}_t\}_{t=1}^T$ comprises local joint rotations $\hat{\theta}_t$, body shape parameters $\hat{\beta}_t$, and per-joint contact probabilities $\hat{\psi}_t$.
For $\hat{\theta}_t$ and $\hat{\beta}_t$,
we utilize the rotation loss 
$
\mathcal{L}_{\text{rot}}
= \frac{1}{T}\sum_{t=1}^T
\| \hat{\theta}_t - \theta_t \|_1
$
and the shape parameter loss
$
\mathcal{L}_{\text{shape}}
= \frac{1}{T}\sum_{t=1}^T
\|\hat{\beta}_t - \beta\|_1
$
, where $\theta_t$ and $\beta$ denote the corresponding ground truth of joint rotations and shape parameters, respectively. 
$T$ indicates the total number of time steps as mentioned.
Note that such joint rotations and body shape parameters are also used to compute the position loss $\mathcal{L}_{\text{pos}}$ as follows:
\begin{equation}
\label{eq:rec}
\mathcal{L}_{\text{pos}}
= \frac{1}{T}\sum_{t=1}^T
\Big\|
\text{FK}\big(\hat{\theta}_{t},\hat{\beta}_t\big)
-
\text{FK}\big(\theta_{t},\beta\big)
\Big\|_{1},
\end{equation}
where \text{FK} denotes forward kinematics of the SMPL-H model~\cite{loper2023smpl}.
In addition, the contact loss $\mathcal{L}_{\text{contact}}$ is applied to supervise per-joint contact probabilities $\hat{\psi}_t$ using the binary cross-entropy loss~\cite{zhang2024proxycap}.
Therefore, the reconstruction loss $\mathcal{L}_{\text{rec}}$ is formulated as follows:
\begin{equation}
\mathcal{L}_{\text{rec}}
= \mathcal{L}_{\text{rot}} + \mathcal{L}_{\text{pos}}
+ \lambda_{\text{shape}}\mathcal{L}_{\text{shape}}
+ \lambda_{\text{contact}}\mathcal{L}_{\text{contact}},
\label{eq:3}
\end{equation}
where weighting factors $\lambda_{\text{shape}}$ and $\lambda_{\text{contact}}$ are both set to 0.003 to balance their scales.
$\mathcal{L}_{\text{KL}}$ is defined as the sum of $\mathcal{L}^{H,M}_{\text{KL}}$ and $\mathcal{L}^{\text{normal}}_{\text{KL}}$, which are computed by using Eqs.~(\ref{eq:1}) and~(\ref{eq:2}).
The velocity loss $\mathcal{L}_{\text{vel}}$ is employed to enforce the temporal smoothness, as follows~\cite{zheng2023realistic}:
\begin{equation}
\label{eq:vel}
\mathcal{L}_{\text{vel}}
=\frac{1}{T-1}\sum_{t=2}^{T}
\Big\|
\big(\hat{p}_{t}-\hat{p}_{t-1}\big)
-
\big(p_{t}-p_{t-1}\big)
\Big\|_{1},
\end{equation}
where \(\hat{p}\) and \(p\) denote the predicted joint position and the ground truth, respectively.
The total loss function is defined as follows:
\begin{equation}
\begin{aligned}
    \mathcal{L}_\text{total} &= \mathcal{L}_\text{rec} + \lambda_\text{KL}\mathcal{L}_\text{KL} + 
    \lambda_\text{vel}\mathcal{L}_\text{vel},
\end{aligned}
\label{eq:3}
\end{equation}
where \(\lambda_\text{KL}\) and \(\lambda_\text{vel}\) are balancing factors, which are empirically set to 0.0004 and 0.003 respectively.

\subsection{Inference for Arbitrary-length Inputs}
\label{sec:arbitrary}
To process sequences of arbitrary length $T$ during the test time, we employ a sliding window strategy following previous approaches~\cite{jiang2022avatarposer, jiang2024egoposer, zheng2023realistic}.
Specifically, for initial $128$ frames, we utilize the full set of predictions generated from a single forward pass.
For subsequent frames, the window slides forward one frame at a time, and the last predicted full-body pose is adopted as the final output.
This strategy is naturally adaptable to an online setting.
The proposed method enables immediate inference by repeatedly filling the window with the first observed head pose, rather than waiting to buffer initial $128$ frames.

\section{Experimental Results}
\label{sec:experimental}

\subsection{Implementation Details}
All experiments were conducted by using the PyTorch framework~\cite{Paszke17torch}, running on an Intel E5-1650 v4@3.60GHz CPU and an NVIDIA RTX 3090 Ti GPU.
We employed the AdamW optimizer~\cite{loshchilov2017decoupled} to train all model parameters, 
with momentum factors \(\beta_1\) and \(\beta_2\) set to 0.9 and 0.999, respectively.
The proposed method is trained for 300 epochs with the batch size of 32.
The learning rate is fixed to \(1\times10^{-4}\) throughout the training process.
Inputs are randomly windowed to a length of $128$ from pairs of head and full-body motion sequences for training.

\subsection{Datasets and Evaluation Metrics}
\noindent\textbf{Dataset.}
For the performance evaluation of the proposed method, two representative benchmarks, i.e., AMASS~\cite{mahmood2019amass} and RICH~\cite{huang2022capturing}, are employed.
We follow the settings used in previous work~\cite{yi2025estimating} with the same split of the AMASS dataset.
For the RICH dataset, we use the test split only for the evaluation as introduced in~\cite{huang2022capturing}.
Given synthetic device poses for both datasets, we adopt the annotation process~\cite{yi2025estimating}, where a central pupil frame is anchored between vertices corresponding to left and right pupils in the mesh.

\begin{table*}[tb]
    \caption{Performance comparisons of ego-body mesh reconstruction based on the AMASS~\cite{mahmood2019amass} dataset (best results and second results are shown in bold and underlined, respectively).}
    \label{tab:1}
    
    \vspace{-1mm}
    
    \centering
    \fontsize{7pt}{9pt}\selectfont
    \setlength{\tabcolsep}{3pt} 
    \begin{tabular}{c cccccc}
        \toprule
        \multirow{2}{*}[-0.8ex]{Methods}
          & \multicolumn{6}{c}{\textbf{AMASS}} \\
        \cmidrule(lr){2-7}
          & MPJPE \(\downarrow\) & PA-MPJPE \(\downarrow\) 
          & Ground \(\downarrow\)
          & \(\textbf{T}_{head}\) \(\downarrow\) 
          & Jitter \(\downarrow\)
          & Foot sliding \(\downarrow\) \\
        \hline\hline
        \addlinespace[1.1pt]
        AvatarPoser$^\dagger$~\cite{jiang2022avatarposer}      
        & 142.2 & 127.5 & 48.1 & 42.9 & 4.74 & 13.5
        \\
        EgoPoser$^\dagger$~\cite{jiang2024egoposer}                           
        & 143.9 & 121.0 & 49.3 & 40.8 & 5.27 & 10.6        
        \\
        EgoEgo~\cite{li2023ego}                           
        & 167.4 & 145.8 & 47.1 & 54.9 & 4.22 & 11.7 
        \\
        EgoAllo~\cite{yi2025estimating}                           
        & \underline{119.7} & \underline{101.1} & \underline{26.3} & \underline{6.2} & \underline{4.06} & \underline{10.2} 
        \\
        \addlinespace[1.1pt]
        \hline 
        \addlinespace[1.1pt]
        EgoGVAE (Ours)         
        & \textbf{106.7} & \textbf{89.9} & \textbf{21.6} & \textbf{5.5} & \textbf{3.08} & \textbf{8.2} \\
        \addlinespace[-1.1pt]
        \bottomrule
        \addlinespace[2pt]
        \multicolumn{7}{l}{$^\dagger$ indicates modified versions of baselines that are retrained using only head pose inputs.}
    \end{tabular}
    \vspace{-2mm}
\end{table*}

\begin{table*}[tb]
    \caption{Performance comparisons of cross-validation based on the RICH~\cite{huang2022capturing} dataset (best results and second results are shown in bold and underlined, respectively). Note that all methods here are trained by using the AMASS~\cite{mahmood2019amass} dataset.
    }
    \label{tab:table_rich}
    
    \vspace{-1mm}
    
    \centering
    \fontsize{7pt}{9pt}\selectfont
    \setlength{\tabcolsep}{3pt} 
    \begin{tabular}{c cccccc}
        \toprule
        \multirow{2}{*}[-0.8ex]{Methods}
          & \multicolumn{6}{c}{\textbf{RICH}} \\
        \cmidrule(lr){2-7}
          & MPJPE \(\downarrow\) & PA-MPJPE \(\downarrow\) 
          & Ground \(\downarrow\) 
          & \(\textbf{T}_{head}\) \(\downarrow\) 
          & Jitter \(\downarrow\)
          & Foot sliding \(\downarrow\)\\
        \hline\hline
        \addlinespace[1.1pt]
        AvatarPoser$^\dagger$~\cite{jiang2022avatarposer}         
        & 297.5 & 287.4 & 281.8 & 75.6 & 4.43 & 12.4 
        \\
        EgoPoser$^\dagger$~\cite{jiang2024egoposer}                           
        & 314.4 & 305.9 & 161.3 & 108.5 & 5.12 & \textbf{9.5} 
        \\
        EgoEgo~\cite{li2023ego}                           
        & 210.7 & 180.1 & 93.6 & 134.5 & \textbf{3.61} & 12.7 
        \\
        EgoAllo~\cite{yi2025estimating}                           
        & \underline{193.1} & \underline{172.9} & \underline{71.9} & \underline{7.7} & 5.07 & 15.9 
        \\
        \addlinespace[1.1pt]
        \hline 
        \addlinespace[1.1pt]
        EgoGVAE (Ours)         
        & \textbf{179.9} & \textbf{165.9} & \textbf{59.0} & \textbf{5.4} & \underline{3.87} & \underline{11.9} \\
        \addlinespace[-1.1pt]
        \bottomrule
        \addlinespace[2pt]
        \multicolumn{7}{l}{$^\dagger$ indicates modified versions of baselines that are retrained using only head pose inputs.}
    \end{tabular}
    \vspace{-2mm}
\end{table*}

\noindent\textbf{Evaluation metrics.}
{\sloppy
For the performance evaluation on 3D human mesh datasets, we use four metrics, i.e., mean per joint position error (MPJPE)~\cite{ionescu2013human3}, Procrustes-aligned mean per joint position error (PA-MPJPE)~\cite{zhou2018monocap}, grounding metric (Ground)~\cite{zheng2023realistic}, and mean head joint position error ($\textbf{T}_\text{head}$)~\cite{yi2025estimating}.
Specifically, MPJPE measures the average value of the Euclidean distance between the predicted 3D joint and the corresponding ground truth.
PA-MPJPE indicates MPJPE calculated after applying the Procrustes analysis to the estimated body mesh.
Note that Ground denotes the average value of the vertical distance between the lowest position of predicted joints and that of ground truth joints to evaluate floating and ground penetration artifacts.
$\textbf{T}_\text{head}$ computes the average value of the Euclidean distance between the predicted position of the head joint and the corresponding ground truth.
\par}

Besides the above pose accuracy, we also use two metrics to assess the quality of the generated motion, i.e., Jitter~\cite{flash1985coordination} and Foot sliding metrics, as defined in~\cite{zheng2023realistic}.
Specifically, Jitter measures the average value of the jerk, i.e., the time derivative of acceleration, over the predicted motion sequence to evaluate the temporal smoothness.
Foot sliding computes the average horizontal displacement of foot joints between consecutive frames to detect sliding artifacts.

\begin{figure*}[tb]
    \centering
    \centerline{\includegraphics[width=1\textwidth]{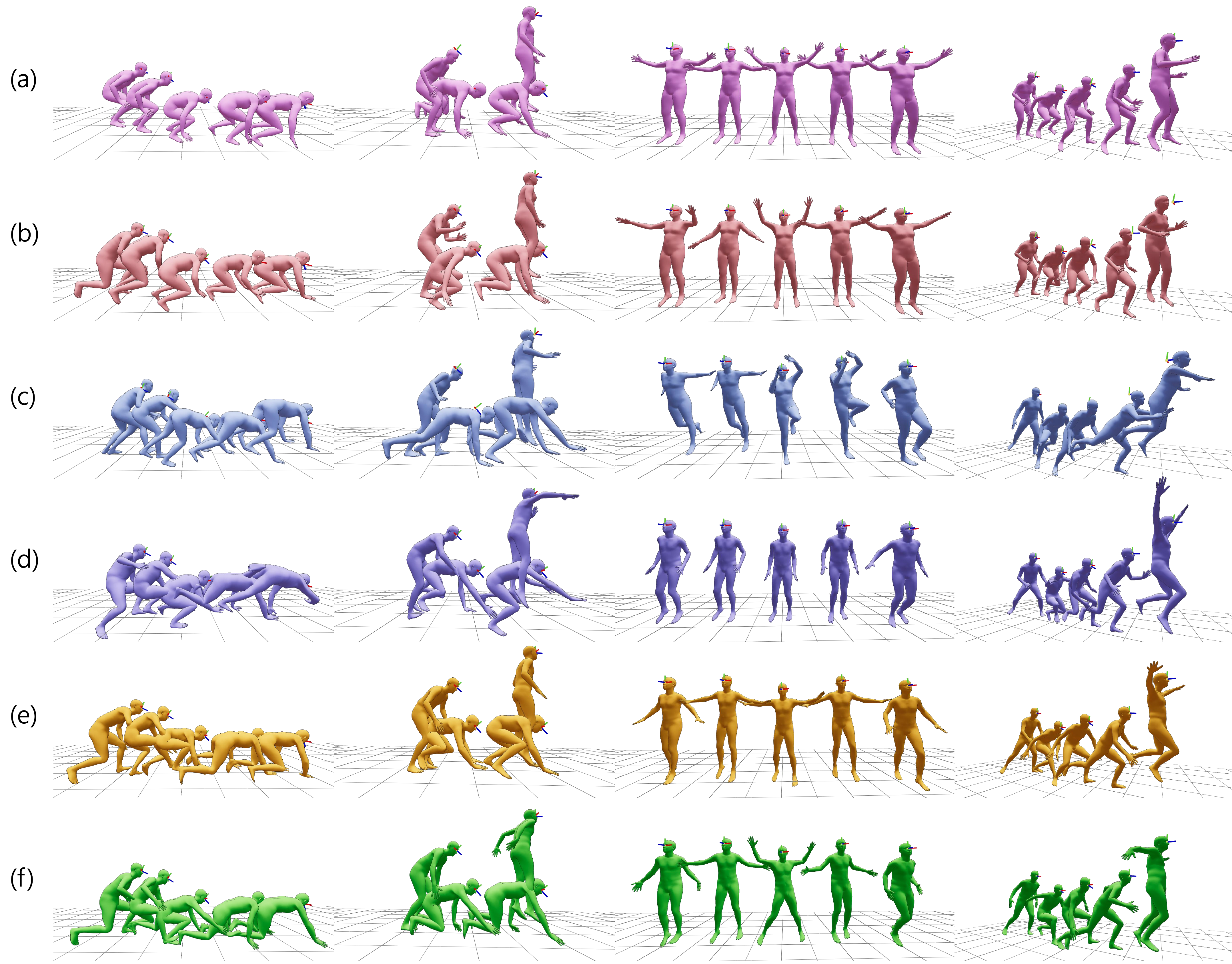}}
    \vspace{-2mm}
    \caption{Results of ego-body mesh reconstruction on the AMASS~\cite{mahmood2019amass} dataset.
    (a) Results by AvatarPoser~\cite{jiang2022avatarposer}. 
    (b) Results by EgoPoser~\cite{jiang2024egoposer}. 
    (c) Results by EgoEgo~\cite{li2023ego}. 
    (d) Results by EgoAllo~\cite{yi2025estimating}. 
    (e) Results by the proposed method (i.e., EgoGVAE). 
    (f) Ground Truth.
    }
    \label{fig:result1} 
    \vspace{-3mm}
\end{figure*}

\begin{figure*}[tb]
    \centering
    \centerline{\includegraphics[width=1\textwidth]{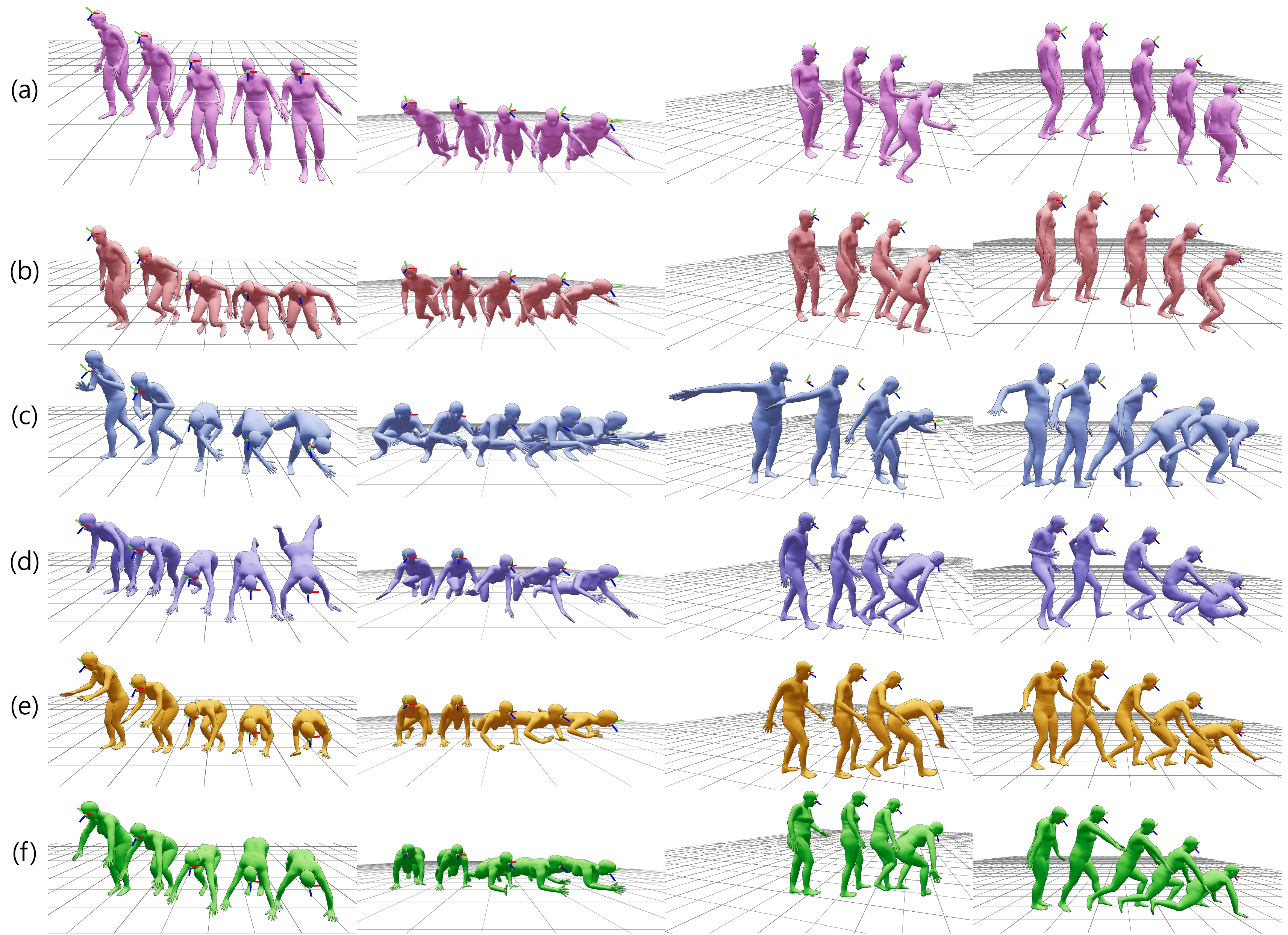}}
    \vspace{-2mm}
    \caption{Results of ego-body mesh reconstruction on the RICH~\cite{huang2022capturing} dataset. 
    (a) Results by AvatarPoser~\cite{jiang2022avatarposer}. 
    (b) Results by EgoPoser~\cite{jiang2024egoposer}. 
    (c) Results by EgoEgo~\cite{li2023ego}. 
    (d) Results by EgoAllo~\cite{yi2025estimating}. 
    (e) Results by the proposed method (i.e., EgoGVAE). 
    (f) Ground Truth.
    }
    \label{fig:result2}
    \vspace{-3mm}
\end{figure*}

\subsection{Performance Evaluation}
\noindent\textbf{Quantitative evaluation.}
{\sloppy
To demonstrate the effectiveness of the proposed method in ego-body mesh reconstruction, we compare ours with previous methods in ego-body mesh reconstruction, i.e., AvatarPoser~\cite{jiang2022avatarposer}, EgoPoser~\cite{jiang2024egoposer}, EgoEgo~\cite{li2023ego}, and EgoAllo~\cite{yi2025estimating}, using input sequences of 128 frames.
Note that AvatarPoser~\cite{jiang2022avatarposer} and EgoPoser~\cite{jiang2024egoposer} utilize both head and hand positions as inputs, which makes them not directly comparable to our head-only setting.
To be fair, we retrained both models using only head pose inputs on the AMASS~\cite{mahmood2019amass} dataset.
As shown in Table~\ref{tab:1}, the proposed method achieves the meaningful performance improvement compared to previous approaches.
Specifically, the proposed method achieves MPJPE of 106.7, PA-MPJPE of 89.9, Ground of 21.6, and $\textbf{T}_{head}$ of 5.5 on the AMASS dataset with the input sequence of 128 frames, which demonstrates $10.9\%$ improvement in MPJPE, $11.1\%$ improvement in PA-MPJPE, $17.9\%$ improvement in Ground, and $11.3\%$ improvement in head position error, respectively.
Moreover, the proposed method also achieves lower Jitter of 3.08 and lower Foot sliding of 8.2 compared to the best-performed method, i.e., EgoAllo~\cite{yi2025estimating}.
Based on results shown in Table~\ref{tab:1}, we could confirm that the proposed guidance scheme is effective to alleviate the ill-posed properties of head-to-motion generation while providing the reliable performance of ego-body mesh reconstruction. 
Furthermore, we conduct cross-validation by training all methods on the AMASS~\cite{mahmood2019amass} dataset and evaluating them on the RICH~\cite{huang2022capturing} dataset.
The result of the performance comparison is shown in Table~\ref{tab:table_rich}.
As can be seen, non-diffusion models, i.e., AvatarPoser~\cite{jiang2022avatarposer} and EgoPoser~\cite{jiang2024egoposer}, suffer from the significant performance drop under this setting.
In contrast, the proposed method still achieves the reliable performance on the benchmark dataset compared to state-of-the-art methods.
Besides, we compare the proposed method with diffusion-based methods, such as EgoEgo~\cite{li2023ego} and EgoAllo~\cite{yi2025estimating}, on the real-world EgoBody~\cite{zhang2022egobody} dataset (see Table~\ref{tab:egobody}). 
This result shows that the proposed method still provides the reliable performance of ego-body mesh reconstruction, even when head pose inputs are captured by real VR/AR devices and contain realistic noise.
\par}

\begin{table}[tb]
    \centering
    \begin{minipage}{0.49\textwidth}
        \centering
        \makeatletter\def\@captype{table}\makeatother 
        \fontsize{7pt}{9pt}\selectfont 
        \setlength{\tabcolsep}{1pt}  
        \caption{Performance comparisons of ego-body mesh reconstruction based on the real-world EgoBody~\cite{zhang2022egobody} dataset (best results and second results are shown in bold and underlined, respectively).
        }
        \label{tab:egobody} 
        \vspace{-3mm}
            \begin{tabular}{c cccc} 
                \toprule
                \multirow{2}{*}[-0.8ex]{Methods} & \multicolumn{3}{c}{EgoBody} \\
                \cmidrule(lr){2-4}
                & MPJPE $\downarrow$ & PA-MPJPE $\downarrow$ & Jitter \(\downarrow\)\\
                \hline\hline
                \addlinespace[1.1pt]
                EgoEgo~\cite{li2023ego}         & 179.9  & 156.0   & \textbf{1.40} \\
                EgoAllo~\cite{yi2025estimating} & \underline{172.9}  & \underline{149.7}   & 1.77 \\
                \addlinespace[1.1pt]
                \hline
                \addlinespace[1.1pt]
                EgoGVAE                 & \textbf{162.9}  & \textbf{140.4}  & \underline{1.42} \\
                \addlinespace[-1.1pt]           
                \bottomrule
            \end{tabular}
    \end{minipage}
    \hfill 
    \begin{minipage}{0.49\textwidth}
        \centering
        \makeatletter\def\@captype{table}\makeatother 
        \fontsize{7pt}{9pt}\selectfont 
        \setlength{\tabcolsep}{0.8pt}   

        \caption{Efficiency comparison with diffusion-based methods (best results and second results are shown in bold and underlined, respectively).
        }
        \label{tab:cost} 
        \vspace{1mm}
            \begin{tabular}{c c c c}
                \toprule
                \multirow{2}{*}[-0.8ex]{Methods} & \multicolumn{3}{c}{128 frames} \\
                \cmidrule(lr){2-4}
                & Params(M)$\downarrow$ & FLOPs(G)$\downarrow$ & Time(sec)$\downarrow$ \\
                \hline\hline
                \addlinespace[1.1pt]
                EgoEgo~\cite{li2023ego} & \textbf{10.97} & 3025.13 & 7.222 \\
                EgoAllo~\cite{yi2025estimating} & 50.45 & \underline{378.16} & \underline{1.510} \\
                \addlinespace[1.1pt]
                \hline 
                \addlinespace[1.1pt]
                EgoGVAE & \underline{13.88} & \textbf{3.45} & \textbf{0.026} \\
                \addlinespace[-1.1pt]
                \bottomrule
            \end{tabular}
    \end{minipage}
    \vspace{-2mm}
\end{table}

\noindent\textbf{Qualitative evaluation.}
Several results of ego-body mesh reconstruction for the AMASS dataset are shown in Fig.~\ref{fig:result1}. 
We can see that the proposed method successfully recovers 3D human mesh only with head trajectories.
Specifically, in the third column of Fig.~\ref{fig:result1}, previous methods generate plausible full-body poses, but they occasionally tend to generate poses with either static arms or arbitrary movements.
In contrast, the proposed method reconstructs plausible arm poses based on the latent distribution guided by full-body poses.
In particular, the proposed method yields robust results even when the significant change in the head trajectory occurs as shown in the second and fourth columns of Fig.~\ref{fig:result1}.
This is further evident in its ability to effectively estimate full-body poses under extreme conditions, e.g., motions during exercise, as shown in Fig.~\ref{fig:result2}.
In particular, previous methods often fail to accurately predict lower-body poses, which leads to the significant performance drop due to effects of foot penetration or floating legs, whereas the proposed method yields plausible full-body poses even under such complicated scenarios.
Based on this, it is thought that our EgoGVAE works robustly for various types of full-body poses without significant distortions of the pose structure.

\noindent\textbf{Efficiency.}
We compare the model efficiency of the proposed method with that of diffusion-based methods, i.e., EgoEgo~\cite{li2023ego} and EgoAllo~\cite{yi2025estimating}, as shown in Table~\ref{tab:cost}. 
We analyze three metrics, i.e., the number of parameters (Params), floating point operations per second (FLOPs), and inference time (Time), which are conducted based on sequences of 128 frames with RTX 3090 Ti GPU. 
As can be seen, diffusion-based approaches perform with high computational costs, however, the proposed method shows a huge advantage in the efficiency of the inference time. 
Specifically, EgoEgo~\cite{li2023ego} has a low number of parameters, i.e., 10.97M, but the iterative process in its diffusion model requires 7.222 seconds and 3025.13G FLOPs to process 128 frames.
The state-of-the-art method, i.e., EgoAllo~\cite{yi2025estimating}, reduces the sampling steps from 1000 to 30, but it significantly increases the number of parameters to 50.45M and still requires 1.510 seconds and 378.16G FLOPs to process 128 frames.
In contrast, the proposed method achieves the lowest FLOPs of 3.45G and the fastest inference time of 0.026 seconds, which is more than 50 times faster compared to diffusion-based methods.

\begin{table}[tb]
    \centering
    \begin{minipage}{0.49\textwidth}
        \centering
        \makeatletter\def\@captype{table}\makeatother 
        \fontsize{7pt}{9pt}\selectfont 
        \setlength{\tabcolsep}{0.5pt}   

        \caption{Performance analysis of the proposed method according to changes in key components of the head-to-motion network on the AMASS~\cite{mahmood2019amass} dataset.
        }
        \label{tab:ab1} 
            \begin{tabular}{c c cc}
                \toprule
                \multirow{2}{*}[-0.8ex]{\textbf{Guidance}} &
                \multirow{2}{*}[-0.8ex]{\begin{tabular}[c]              {@{}c@{}}\textbf{Learnable}\\\textbf{tokens}\end{tabular}} &
                \multicolumn{2}{c}{\textbf{AMASS}}\\
                \cmidrule(lr){3-4}
                 &  & MPJPE $\downarrow$ & PA-MPJPE $\downarrow$ \\
                \midrule
                \ding{55}  & \ding{55}  & 128.3 & 110.3 \\
                \ding{55}  & \checkmark & 125.6 & 108.6 \\
                \checkmark & \ding{55}  & 113.7 & 94.5 \\
                \checkmark & \checkmark & 106.7 & 89.9 \\
                \bottomrule
            \end{tabular}
    \end{minipage}
    \hfill 
    \begin{minipage}{0.49\textwidth}
        \centering
        \makeatletter\def\@captype{table}\makeatother 
        \fontsize{7pt}{9pt}\selectfont 
        \setlength{\tabcolsep}{4.5pt}   

        \caption{Performance analysis of the proposed method according to different combinations of loss terms on the AMASS~\cite{mahmood2019amass} dataset.
        }
        \label{tab:ab2} 
            \begin{tabular}{c c cc} 
            \toprule
            \multirow{2}{*}[-0.8ex]{\textbf{$\mathcal{L}^{normal}_{KL}$}} & 
            \multirow{2}{*}[-0.8ex]{\textbf{$\mathcal{L}_{vel}$}} &
            \multicolumn{2}{c}{\textbf{AMASS}} \\ 
            \cmidrule(lr){3-4} 
            &  & MPJPE $\downarrow$ & PA-MPJPE $\downarrow$ \\ 
            \midrule
            \ding{55} & \ding{55}  & 116.5 & 99.4 \\ 
            \ding{55} & \checkmark & 114.6 & 97.9 \\ 
            \checkmark & \ding{55}  & 109.8 & 91.8 \\ 
            \checkmark & \checkmark & 106.7 & 89.9 \\ 
            \bottomrule
            \end{tabular}
    \end{minipage}
    \vspace{-2mm}
\end{table}

\begin{figure}[tb]
    \centering
    \centerline{\includegraphics[width=1\textwidth]{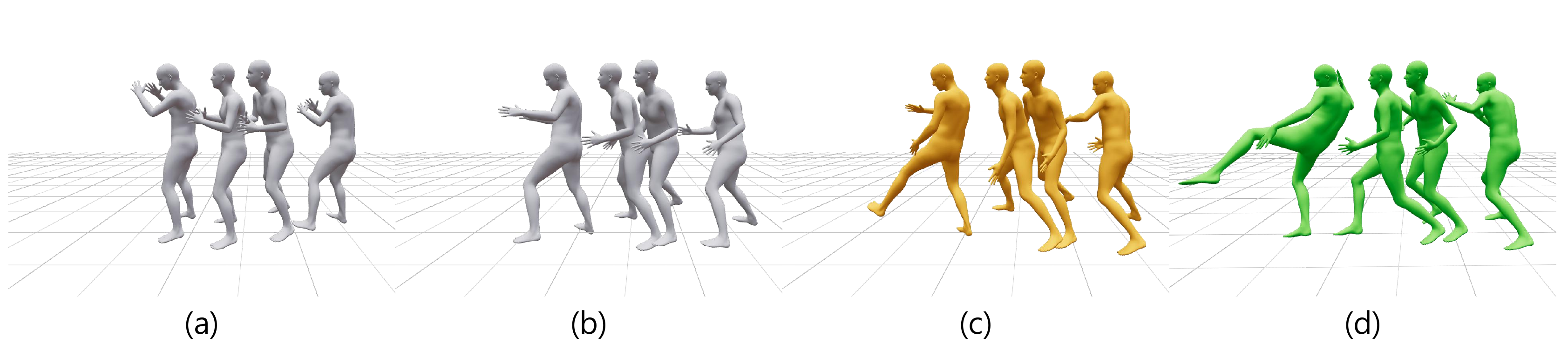}}
    \vspace{-2mm}
    \caption{(a) Result by removing both guidance and learnable tokens, i.e., baseline. (b) Result by baseline $+$ guidance. (c) Result by baseline $+$ guidance $+$ learnable tokens (our method). (d) Ground Truth.
    }
    \label{fig:ab1}
    \vspace{-2mm}
\end{figure}

\subsection{Ablation Study}
In this subsection, we conduct comparative experiments on the AMASS~\cite{mahmood2019amass} dataset to verify the effectiveness of the proposed method.
Specifically, we analyze the impact of key components on ego-body mesh reconstruction by systematically evaluating performance variations of our model.

\noindent \textbf{Effect of EgoGVAE.}
First of all, we check the effect of the guidance by the motion-to-motion network in a variational way. 
As can be seen in Table~\ref{tab:ab1}, it is easy to see that enforcing the latent distribution of the head-to-motion network to be similar to that of the motion-to-motion network significantly improves the reconstruction accuracy (128.3 $\xrightarrow{}$ 113.7 in MPJPE and 110.3 $\xrightarrow{}$ 94.5 in PA-MPJPE).
In what follows, we also check the effect of learnable tokens, which are appended to input tokens for consideration of other body parts except the head. 
From the result shown in Table~\ref{tab:ab1}, we can see that this appended input is helpful for ego-body mesh reconstruction rather than simply using the head token.
By using these two components together, i.e., the proposed method, the best performance can be achieved. We also provide the visual comparison in Fig.~\ref{fig:ab1}.
As can be seen, the lower-body motions are not accurately reconstructed without guidance and learnable tokens (see Fig.~\ref{fig:ab1}(a)).
By adopting the guidance, the dynamics of the lower-body motions has been increased as shown in Fig.~\ref{fig:ab1}(b).
Finally, full-body poses become more flexible and natural with learnable tokens (see Fig.~\ref{fig:ab1}(c)).

\noindent \textbf{Effect of loss terms.}
We analyze the effect of each loss term and the corresponding result is shown in Table~\ref{tab:ab2}.
Note that core loss terms, i.e., the reconstruction loss $\mathcal{L}_{rec}$ and the symmetric Kullback-Leibler divergence $\mathcal{L}^{H,M}_{KL}$, are applied to all the experiments for the stability of the learning process.
As shown in Table~\ref{tab:ab2}, the performance is significantly dropped by removing $\mathcal{L}^{normal}_{KL}$.
This means that regularization of latent distributions plays an important role in our guidance-based learning scheme.
Furthermore, we can see that removing $\mathcal{L}_{vel}$ results in the degradation of 3.1 in MPJPE and 1.9 in PA-MPJPE, respectively. 
This analysis confirms that both loss terms are essential for the best performance.

\noindent \textbf{Effect of network designs.}
We investigate different network designs for guiding the head-to-motion network with the motion-to-motion network, and the corresponding results are shown in Table~\ref{tab:net_design}.
The first row presents the result of the variant where a mapping between the latent space of the head-to-motion network and that of the motion-to-motion network is learned instead of aligning latent distributions of two networks.
The second row shows the result of using a pre-trained motion-to-motion network as a fixed motion prior without joint training.
As can be seen, both variants exhibit the performance degradation across all metrics compared to our default setting. 
This result demonstrates that aligning latent distributions of two networks through joint training is effective for guiding the process of head-to-motion generation.

\begin{table}[t]
    \caption{Performance analysis of the proposed method according to different network designs for guiding the head-to-motion network on the AMASS~\cite{mahmood2019amass} dataset.}
    \label{tab:net_design}
    \vspace{-1mm}
    \centering
    \fontsize{7pt}{9pt}\selectfont
    \setlength{\tabcolsep}{2pt}
    \begin{tabular}{c cccc} 
        \toprule
        \addlinespace[1.5pt]           
        \multicolumn{1}{c}{Methods} 
            & \multicolumn{1}{c}{MPJPE $\downarrow$} 
            & \multicolumn{1}{c}{PA-MPJPE $\downarrow$} 
            & \multicolumn{1}{c}{Ground \(\downarrow\)} 
            & \multicolumn{1}{c}{Jitter $\downarrow$} \\
        \hline\hline
        \addlinespace[1.1pt]
        Ours (learning a latent-to-latent mapping)     & 128.1     & 111.9   & 40.4     & 3.26\\
        Ours (using a pre-trained motion prior)     & 111.2     & 94.5    & 23.9     & 3.19\\
        \addlinespace[1.1pt]
        \hline 
        \addlinespace[1.1pt]
        Ours (default setting)              & 106.7     & 89.9    & 21.6    & 3.08\\
        \addlinespace[-1.1pt]        
        \bottomrule
    \end{tabular}
    \vspace{-2mm}
\end{table}

\begin{table}[tb]
    \centering
    \begin{minipage}{0.49\textwidth}
        \centering
        \makeatletter\def\@captype{table}\makeatother 
        \fontsize{7pt}{9pt}\selectfont 
        \setlength{\tabcolsep}{1pt}   

        \caption{Performance comparisons based on the AMASS~\cite{mahmood2019amass} dataset following the inference setting of longer sequences (best results and second results are shown in bold and underlined, respectively).
        }
        \label{tab:full_1} 
            \begin{tabular}{c ccc}
                \toprule
                \multirow{2}{*}[-0.8ex]{Methods}
                  & \multicolumn{3}{c}{256 frames} \\
                \cmidrule(lr){2-4}
                  & MPJPE \(\downarrow\) & PA-MPJPE \(\downarrow\) 
                  & Jitter \(\downarrow\) \\
                \hline\hline
                \addlinespace[1.1pt]
                EgoEgo~\cite{li2023ego}                           
                & 140.8 & 105.5 & \underline{4.94}
                \\
                EgoAllo~\cite{yi2025estimating}                           
                & \underline{113.4} & \underline{91.8} & 5.29
                \\
                \addlinespace[1.1pt]
                \hline 
                \addlinespace[1.1pt]
                EgoGVAE          
                & \textbf{95.8} & \textbf{73.9} & \textbf{4.54} 
                \\
                \addlinespace[-1.1pt]
                \bottomrule
            \end{tabular}
    \end{minipage}
    \hfill 
    \begin{minipage}{0.49\textwidth}
        \centering
        \makeatletter\def\@captype{table}\makeatother 
        \fontsize{7pt}{9pt}\selectfont 
        \setlength{\tabcolsep}{1pt}   

        \caption{Performance comparisons based on the AMASS~\cite{mahmood2019amass} dataset following the online inference setting (best results and second results are shown in bold and underlined, respectively).
        }
        \label{tab:full_2} 
            \begin{tabular}{c ccc}
                \toprule
                \multirow{2}{*}[-0.8ex]{Methods}
                  & \multicolumn{3}{c}{Online setting} \\
                \cmidrule(lr){2-4}
                  & MPJPE \(\downarrow\) & Jitter \(\downarrow\) 
                  & Time (ms) \(\downarrow\) \\
                \hline\hline
                \addlinespace[1.1pt]
                AvatarPoser~\cite{jiang2022avatarposer}                           
                & \underline{142.2} & \underline{4.74} & \textbf{1.5}
                \\
                EgoPoser~\cite{jiang2024egoposer}                           
                & 143.9 & 5.27 & \underline{1.7}
                \\
                \addlinespace[1.1pt]
                \hline 
                \addlinespace[1.1pt]
                EgoGVAE        
                & \textbf{119.3} & \textbf{4.63} & 26
                \\
                \addlinespace[-1.1pt]
                \bottomrule
            \end{tabular}
    \end{minipage}
    \vspace{-2mm}
\end{table}

\subsection{Quantitative Evaluation on Extended Settings}
\noindent\textbf{Longer sequence evaluation.}
During test time, we adopt the sliding window strategy described in subsection~\ref{sec:arbitrary} to process longer sequences.
To verify stability in this setting, we compare ours with diffusion-based methods, i.e., EgoEgo~\cite{li2023ego} and EgoAllo~\cite{yi2025estimating}, using input sequences of 256 frames.
Specifically, we conduct this evaluation by filtering out sequences in the AMASS~\cite{mahmood2019amass} dataset shorter than 256 frames.
Note that baselines follow the protocol of EgoAllo~\cite{yi2025estimating} to process longer sequences.
As shown in Table~\ref{tab:full_1}, the proposed method achieves the best performance across all metrics, notably the lowest Jitter of 4.54.
This result could confirm that our learned latent distribution ensures the temporal consistency beyond the training length.

\noindent\textbf{Online evaluation.}
To further validate the real-time performance, we compare ours with non-diffusion baselines, i.e., AvatarPoser~\cite{jiang2022avatarposer} and EgoPoser~\cite{jiang2024egoposer}, following the inference process described in subsection~\ref{sec:arbitrary}.
For this evaluation, we utilize the AMASS~\cite{mahmood2019amass} dataset without cropping sequences into fixed-length subsequences.
The result of the performance comparison is shown in Table~\ref{tab:full_2}.
While previous methods achieve faster inference compared to our method, they exhibit notably lower reconstruction accuracy.
In contrast, the proposed method achieves the lowest MPJPE of 119.3 and the lowest Jitter of 4.63, with an inference time of 26 milliseconds per frame, which is still enough for real-time applications.

\section{Conclusion}
\label{sec:conclusion}
We propose a simple yet powerful method for full-body mesh reconstruction from only the head pose of the wearer.
The key idea of the proposed method is to leverage the latent space of the motion-to-motion network, which is a variational autoencoder that takes full-body poses as inputs, to guide the head-to-motion network.
By enforcing two latent distributions, which are encoded from this guidance network and the head-to-motion network respectively, to be similar, 
the proposed method can produce natural full-body poses given the head pose.
Experimental results on benchmark datasets show that our EgoGVAE reliably conducts ego-body mesh reconstruction even with complicated scenarios.
Moreover, the proposed method works very fast compared to previous methods while showing the significant performance improvement.


\section*{Acknowledgements}
This work was supported by the National Research Foundation of Korea (NRF) grant funded by the Korea government (MSIT) (RS-2026-25471545).


%
%
\bibliographystyle{splncs04}
\bibliography{main}
\end{document}